\PassOptionsToPackage{table}{xcolor}
\documentclass[sigconf, screen, nonacm]{acmart}
\usepackage{xcolor}
\usepackage{colortbl}
\usepackage{array}

\acmSubmissionID{4976}

\begin{document}

\title{Make Your MoVe: Make Your 3D Contents by Adapting Multi-View Diffusion Models to External Editing}

\author{Weitao Wang}
\affiliation{
    Tsinghua University
    \country{}
%
}

\author{Haoran Xu}
\affiliation{
    Zhejiang University
    \country{}
%
}

\author{Jun Meng}
\affiliation{
    Zhejiang University
    \country{}
%
}
\authornote{Corresponding author.}

\author{Haoqian Wang}
\affiliation{
    Tsinghua University
    \country{}
%
}
\authornotemark[1]


\begin{abstract}
  As 3D generation techniques continue to flourish, the demand for generating personalized content is rapidly rising. Users increasingly seek to apply various editing methods to polish generated 3D content, aiming to enhance its color, style, and lighting without compromising the underlying geometry. However, most existing editing tools focus on the 2D domain, and directly feeding their results into 3D generation methods (like multi-view diffusion models) will introduce information loss, degrading the quality of the final 3D assets. In this paper, we propose a tuning-free, plug-and-play scheme that aligns edited assets with their original geometry in a single inference run. Central to our approach is a geometry preservation module that guides the edited multi-view generation with original input normal latents. Besides, an injection switcher is proposed to deliberately control the supervision extent of the original normals, ensuring the alignment between the edited color and normal views. Extensive experiments show that our method consistently improves both the multi-view consistency and mesh quality of edited 3D assets, across multiple combinations of multi-view diffusion models and editing methods.
\end{abstract}


\begin{teaserfigure}
  \includegraphics[width=\textwidth]{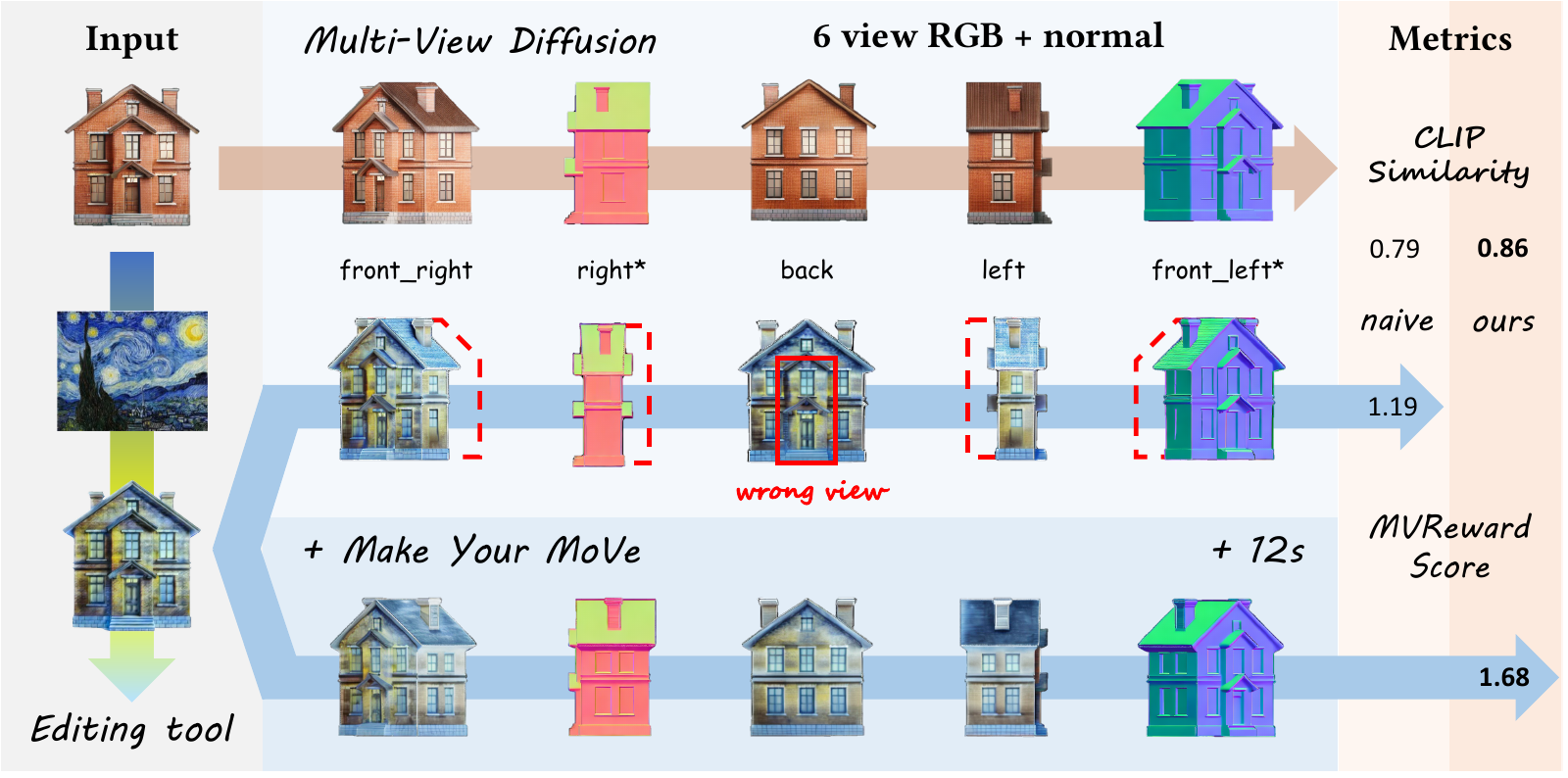}
  \caption{The \emph{naive} approach of simply feeding the edited image into multi-view diffusion models may lead to geometric inaccuracies across views and wrong texture in the back view (view* means its normal map), owing to information blurring introduced by external editing tools. Make Your MoVe addresses this issue through one inference process, achieving better results in metrics (CLIP Similarity is calculated across the generated views and MVReward Score reflects human preferences.}
  \label{fig:teaser}
\end{teaserfigure}

\settopmatter{printacmref=false}

\maketitle

\section{Introduction}
\begin{figure*}[t]
\centering
\includegraphics[width=0.9\textwidth]{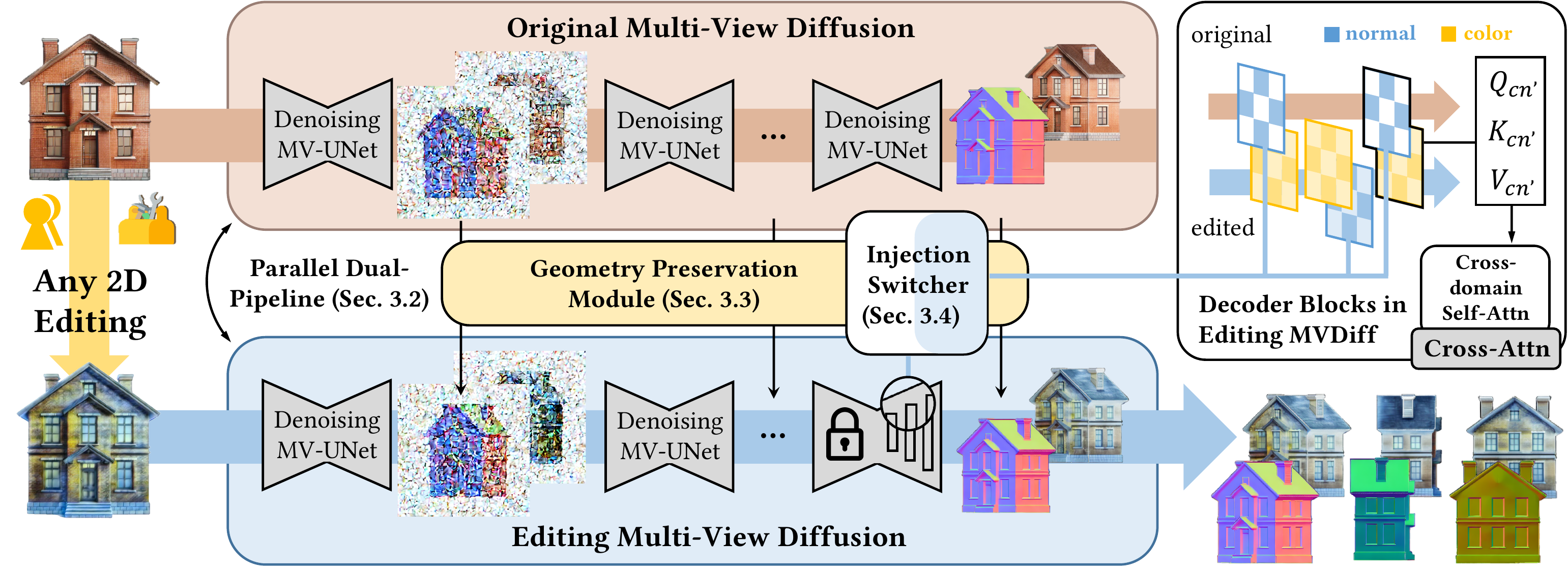} 
\caption{An overview of our whole framework. We first construct a parallel dual-pipeline, which takes both the original image and the user-edited image as inputs (Sec 3.2). The multi-view diffusion model and the 2D black-box editing tool can be any possible combination specified by the user. Next, during a single inference, we present a geometry preservation module to supervise the edited pipeline generation with original normals (Sec 3.3). To address potential issues with color and normal misalignment, we design an injection switcher, which alternately switches between normal latents, allowing the edited color latents to be periodically combined with both the original and edited normals in the cross-domain self-attention (Sec 3.4).}
\label{pipe}
\vspace{-1em}
\end{figure*}

When Van Gogh was painting \emph{The Starry Night} in 1889, he could hardly imagine that, more than a century later, people would be extracting the painterly strokes from his work using style transfer techniques. While the interplay of blue and yellow across the sky and the dark green silhouettes of trees can be captured, there remains one critical element missing: the spatial perspective formed by nearby trees, distant houses, and the starry sky, which grants the painting a flowing and spatial sensation not found in standard 2D images. Extending the stylistic edited 2D images into a higher dimension—a task now emerging as a new frontier—is posing challenges and gathering attention. Beyond style transfer, other 2D appearance editing approaches like prompt instruction and relighting also await adaptation into the 3D domain for further applications.

A straightforward but naive attempt might be simply feeding the edited 2D images into existing 3D generation methods, such as multi-view diffusion models, to generate other viewpoints with corresponding editing. However, as fully discussed in \cite{stylizenew, 3dsty, stylerf}, and illustrated in Figure \ref{fig:teaser}, external edits can introduce some information loss (e.g., style injection might obscure underlying geometric information), resulting in blurred or viewpoint-inconsistent 3D outputs. This highlights the need for a solution that can adapt multi-view diffusion models to external edits, preserving both the quality of the original 3D geometry and the intended edit in texture.

In this paper, we present a novel tuning-free scheme that integrates seamlessly into a single inference run—polishing the edited object’s structure within seconds. Our method builds on pre-trained multi-view diffusion models and 2D black-box editing tools (style transfer, prompt instruction, relighting, ControlNet\cite{controlnet}-based editing, and so on), serving as a plug-and-play strategy compatible with any combination of these methods. 

Specifically, we first define the task objective as generating 3D assets from edited images while preserving the original geometry and allowing for texture changes. Considering that the normal maps in multi-view diffusion models primarily reflect geometric structure, we propose aligning the normal maps between the pre- and post-editing stages while retaining changes in the color images. To achieve this, we first construct a parallel dual-pipeline with both the original and edited inputs. Then a geometry preservation module is designed to replace the normal latents of the edited input with those of the original input at each inference step, ensuring the original geometric supervision. While directly replacing the normal latents may lead to misalignments between the color and normal outputs in our observation, we further introduce an injection switcher to address this issue by alternately switching between the original and edited normals under the user's control, then combining with the edited colors in cross-domain self-attention. Figure \ref{pipe} clearly overviews our whole framework.

To validate the superiority and generalizability of our approach, we test on two commonly used multi-view diffusion models with different output resolutions (256$^2$ and 512$^2$), Wonder3D\cite{wonder3d} and Era3D\cite{era3d}, combining each with four editing methods—StyleTR$^2$\cite{stytr2} and StyleShot\cite{styleshot} (style transfer), InstructPix2Pix\cite{instructpix2pix} (prompt instruction) and IC-Light\cite{iclight} (relighting). We also compare with a SOTA method (Style3D\cite{style3d}) in 3D style transfer based on multi-view images. Experiments show that our method consistently improves the 3D generation quality of edited images, as demonstrated in both multi-view images and mesh reconstructions.

Our main contributions are threefold:
\begin{itemize}
    \item We are the first to propose a tuning-free scheme that adapts multi-view diffusion models to external edits, offering a superior alternative to naively combining an 2D edited input with multi-view diffusion.
    \item We design a novel geometry preservation module with an injection switcher to align the edited geometry with original one with retained texture changes. As a plug-and-play solution, it impressively enhances the 3D editing performance of pre-trained models within seconds.
    \item Extensive experiments demonstrate that our approach robustly improves the multi-view and mesh quality of edited 3D assets across multiple combinations of multi-view diffusion models and editing methods.
\end{itemize}

\section{Related Work}

\subsection{3D Generation with Multi-View Diffusion}

3D generation is advancing rapidly nowadays, with DreamFusion\cite{dreamfusion} setting a precedent. However, distillation-based methods\cite{prolificdreamer, sjc}, following DreamFusion introduce Score Distillation Sampling (SDS) loss along with time-consuming per-shape optimization and multi-head problems that need to be addressed. Meanwhile, direct-3D methods\cite{instant3d, instantmesh, lrm, grm, lgm} are constrained by the limited scale of publicly available 3D asset datasets. As a compromise, multi-view diffusion models \cite{zero, syncdreamer, v3d, sv3d} are receiving increasing attention.

Zero123\cite{zero} has pioneered this field by training a diffusion model conditioned on the reference image and camera viewpoint, which can generate novel views of the same object. Zero123++\cite{zero123++} enhances this ability by improving multi-view consistency and image conditioning. Wonder3D\cite{wonder3d} designs a cross-domain attention mechanism to bring normal maps into multi-view generation. Envision3D\cite{envision3d} increases the number of the generated multi-view images by adopting a two-stage training strategy. Era3D\cite{era3d} can generate high-resolution multi-view images without shape distortions by proposing a row-wise attention layer and diffusion-based camera prediction module. Our work focuses on adapting these powerful generative models to external editing due to their view-consistent results, high compatibility with downstream tasks, and promising development vision.

\subsection{Appearance Editing with Preserved Content}

Appearance editing refers to the transformation or enhancement of the visual style and aesthetic elements of a visual media (e.g., images, videos, 3D assets) without altering its original content. This process is also classified as stylistic editing, as discussed in the editing method survey\cite{editsurvey}, where illumination editing (relighting) and some prompt instruction editing also fit this definition.  

Image style transfer \cite{styleclip, clipstyler, inst} aims to combine the content of one image with the visual style of another, often leveraging deep neural networks such as GANs\cite{gan} and AEs\cite{ae} to extract and inject style features. The capabilities of large text-to-image (T2I) models (like Imagen\cite{imagen} and Stable Diffusion\cite{ldm}), are also being explored by style-tuning methods\cite{specialist, diffinstyle}. Image relighting \cite{stylitgan, diffusionlight, lightit} is another form of appearance editing, involving altering an image’s lighting conditions to simulate different illumination settings. Recent advances\cite{relightful, iclight} utilize deep learning, including diffusion models, to predict and modify lighting effects. Both style transfer and relighting have been extended into the 3D domain. Examples include StyleRF\cite{stylerf} and StyleGaussian\cite{stylegaussian} for 3D style transfer, with IllumiNeRF\cite{illuminerf} and GS$^3$\cite{gs3} for 3D relighting. However, most of these methods focus on NeRF\cite{nerf} or 3DGS\cite{3dgs} representations, overlooking the migration to multi-view images and mesh.

The most related concurrent work to ours is Style3D\cite{style3d}, which focuses on achieving 3D object stylization from the content-style pair, instead of using existing 2D style transfer tools. However, it does not include a comparison with straightforward combinations of style transfer and multi-view diffusion model, a necessary discussion that we focus on in this paper. In terms of extracting and preserving original content, StyleShot\cite{styleshot} uses contouring and line-arting as 2D supervisory signals, whereas we employ normal maps from multiple viewpoints for 3D supervision.

\subsection{Fine-tuning and Tuning-free (PnP) Personalization of Diffusion Models}

With the powerful generative capabilities of diffusion models, several attempts to personalize them are emerging, broadly categorized into fine-tuning and tuning-free approaches. Fine-tuning methods like \cite{dreambooth, tinv, animatediff, dreambooth3d} require a collection of images for user-specified subjects or styles to fine-tune the pre-trained model for personalization. In contrast, tuning-free methods \cite{pnp, styleid, style3d, anyv2v} often leverage pre-trained multimodal encoders to project personalized content into learned latent spaces or directly manipulate intermediate features of attention layers or image latents to inject personalized editing, of which the latter is often referred to as Plug-and-Play (PnP) methods. StyleID\cite{styleid} adapts the pre-trained diffusion model to the image style transfer task by manipulating self-attention features. AnyV2V\cite{anyv2v} combines 2D image editing tools with video diffusion models through a plug-and-play feature injection framework.

Inspired by the tuning-free, especially PnP methods above, our method controls the intermediate variables during a single inference and achieves seamless personalization of multi-view diffusion models without any optimization, leveraging existing editing tools.

\section{Method}

\subsection{Preliminaries}
\textbf{Diffusion Models} \cite{ddpm, ddim, ldm} are a class of generative models that 
(1)~transform data into noise through a \textbf{forward diffusion process} based on the idea of Markov chain\cite{markov}:
\begin{equation}
\begin{aligned}
q(\mathbf{x}_t | \mathbf{x}_{t-1}) &= \mathcal{N}\left(\mathbf{x}_t; \sqrt{1 - \beta_t} \, \mathbf{x}_{t-1}, \beta_t \mathbf{I}\right),\\
q(\mathbf{x}_t | \mathbf{x}_0) &= \mathcal{N}\left(\mathbf{x}_t; \sqrt{\bar{\alpha}_t} \, \mathbf{x}_0, \left(1 - \bar{\alpha}_t\right) \mathbf{I}\right),
\end{aligned}
\end{equation}
where \( \beta_t \) controls the noise variance at each step, \( \alpha_t = 1 - \beta_t \) and \( \bar{\alpha}_t = \prod_{s=1}^t \alpha_s \), indicating \( \mathbf{x}_t \) retains a scaled version of \( \mathbf{x}_0 \) plus accumulated noise.

\noindent(2)~and then recover the data via a \textbf{reverse denoising process} leveraging a neural network\cite{unet, transformer}:
\begin{equation}
p_{\theta}(\mathbf{x}_{t-1} | \mathbf{x}_t) \approx \mathcal{N}\left(\mathbf{x}_{t-1}; \mu_{\theta}(\mathbf{x}_t, t), \Sigma_{\theta}(\mathbf{x}_t, t)\right),
\end{equation}
where $\theta$ are the learnable parameters of the neural network, with \( \mu_{\theta}(\mathbf{x}_t, t) \) and \( \Sigma_{\theta}(\mathbf{x}_t, t) \) predicting the mean and variance.

The goal of the whole training process is to minimize the discrepancy between the predicted noise and the added noise across all time steps, ensuring effective reversal of the diffusion process:
\begin{equation}
\mathcal{L}(\theta) = \mathbb{E}_{q(\mathbf{x}_0)} \left[ \sum_{t=1}^T \mathbb{E}_{q(\mathbf{x}_t | \mathbf{x}_0)} \left[ \left\| \mathbf{\epsilon}_\theta(\mathbf{x}_t, t) - \mathbf{\epsilon}_t \right\|^2 \right] \right].
\end{equation}
where \( \mathbf{\epsilon}_t = \mathbf{x}_t - \sqrt{\bar{\alpha}_t} \, \mathbf{x}_0 \) is the noise added at step \( t \), and \( \mathbf{\epsilon}_\theta(\mathbf{x}_t, t) \) is the model's prediction. Minimizing this loss enables the model to accurately predict and remove noise during the reverse process, thus learning to generate high-fidelity samples.

\noindent\textbf{Multi-view Diffusion Models} \cite{zero, wonder3d, era3d} extend 2D diffusion models to multiple novel view generation. Traditional approaches to representing 3D assets often rely on formats such as point clouds \cite{pointe}, neural radiance fields \cite{nerf}, or tri-planes \cite{eg3d}, and so on. In contrast, Wonder3D \cite{wonder3d} proposes a distinctive method that models the distribution of 3D assets, \(p_a(z)\), as a joint probability distribution over 2D multi-view normal maps and color images. Given a set of camera views \(\{\pi_1, \pi_2, \dots, \pi_K\}\) and a conditional input image \(y\), this distribution can be expressed as:
\begin{equation}
    p_a(z) = p_{nc}(x^{1:K}, n^{1:K} \mid y),
\end{equation}
where the distribution of the color images \(x^{1:K}\) and normal maps \(n^{1:K}\) is observed as \(p_{nc}\). The objective of this work is to design a model \(g_\theta\) that generates color images and normal maps under a given input \(y\) and camera configurations \(\pi_{1:K}\):
\begin{equation}
    (n^{1:K}, x^{1:K}) = g_\theta(y, \pi_{1:K}).
\end{equation}
Overall, the joint distribution of normal maps and color images can be formulated as a Markov chain in the context of the diffusion framework with Gaussian noises \(p(n^{1:K}_T, x^{1:K}_T)\):
\begin{equation}
    p(n^{1:K}_T, x^{1:K}_T) \prod_t p_\theta(n^{1:K}_{t-1}, x^{1:K}_{t-1} \mid n^{1:K}_t, x^{1:K}_t),
\end{equation}

\subsection{Task Formulation}

Our target application scenario is one in which the user applies external, personalized edits to the original image, and expects the multi-view diffusion output to reflect changes in color, style, and aesthetics (i.e., texture in 3D) while preserving the underlying geometric structure. 

As discussed in the introduction, directly feeding the edited image into multi-view diffusion may cause the appearance edits to overwrite or obscure the original geometric information. This leads to discrepancies in the normal maps and ultimately produces undesired geometric distortions in the final 3D asset (mesh). Other simple combinations, such as first generating multiple views from the original input and then applying style transfer to each output individually, also can result in inconsistent viewpoints.

We then formulate the problem as follows: given an original image $I_o$ and a pre-trained 2D editing tool $T$, we can obtain the edited 2D image $I_e$ = $T(I_o)$. Different from other 3D editing methods \cite{stylizenew, style3d}, we focus on preserving geometric consistency and maintaining alignment across views during the editing process. Thus, we directly leverage the pre-trained 2D editing methods, without the intricate and time-consuming learning of editing operations (such as style feature extraction and injection) from scratch in the 3D domain.

The multi-view diffusion model is employed as $g_\theta$, with a pre-trained parameter $\theta$. When provided with an input image $I$, the model produces multi-view representations that include color images and normal maps from six different viewpoints:
\begin{equation}
    g_\theta(I) = \{ (C_i(I), N_i(I)) \}_{i=1}^6,
\end{equation}

where $C_i(I)$ is the color image and $N_i(I)$ is the normal map corresponding to the $i$-th viewpoint. For more details on the mathematical preliminaries of diffusion models and multi-view diffusion models, refer to Appendix 3.

To retain geometric consistency before and after the external editing, we first need the pre-edit results as prior knowledge. Inspired by \cite{controlnet, animate, anyv2v} using an additional network to provide conditional supervising signals in their respective fields, we construct a \textbf{parallel dual-pipeline} sharing the same $g_\theta$:
\begin{equation}
\begin{aligned}
    g_\theta^o &= g_\theta(I_o) = \{ (C_i(I_o), N_i(I_o)) \}_{i=1}^6 \\
    g_\theta^e &= g_\theta(I_e) = \{ (C_i(I_e), N_i(I_e)) \}_{i=1}^6
\end{aligned}
\label{eq2}
\end{equation}

$g_\theta(I_o)$ is referred to as the original pipeline and $g_\theta(I_e)$ as the edited pipeline. The outputs of $g_\theta(I_o)$ serve as supervisions for the geometric structure, while $g_\theta(I_e)$ contains the edited appearance information we wish to incorporate.

\begin{figure}[t]
\centering
\includegraphics[width=0.48\textwidth]{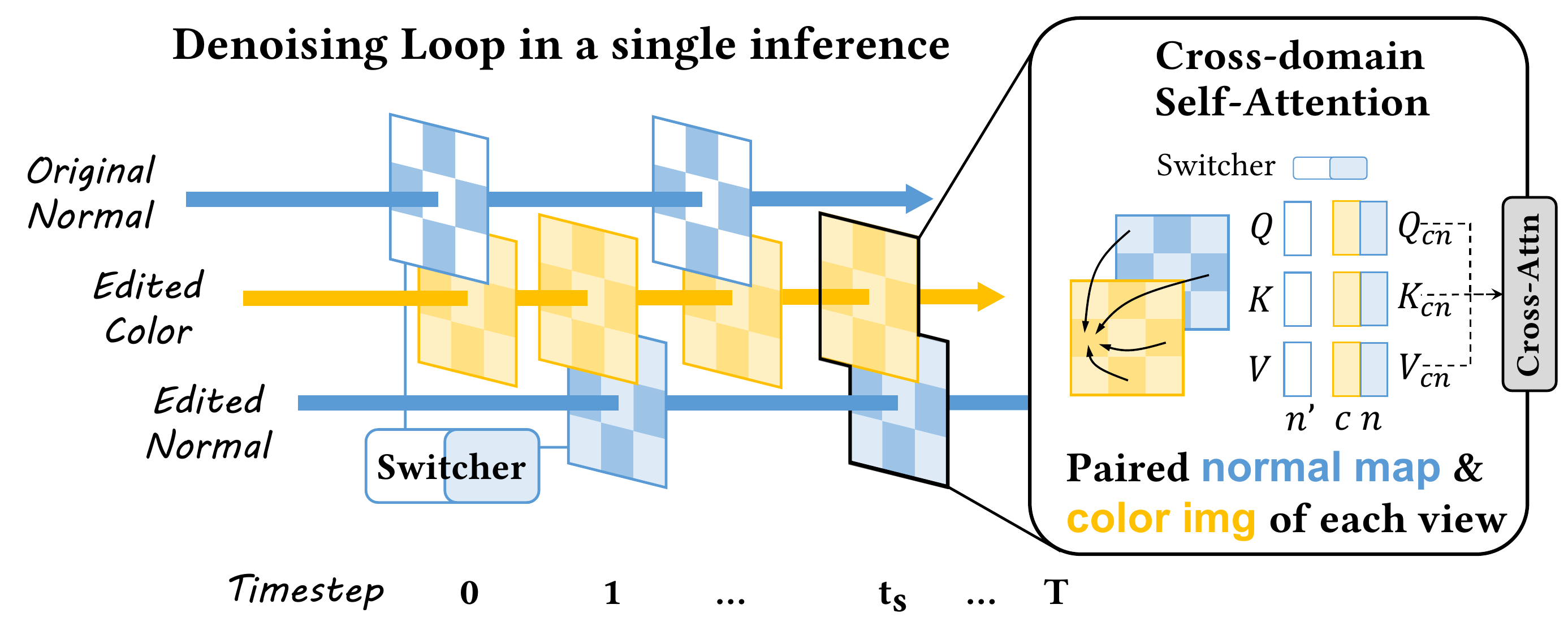} 
\caption{Specific illustration of our normal preservation module and injection switcher in cross-domain self-attention.}
\label{archi}
\end{figure}

\subsection{Geometry Preservation Module}

Starting from our goal of preserving the geometry of edited assets, we argue that the geometric information of multi-view images is mainly embedded in the normal maps, as similarly asserted by \cite{wonder3d, era3d}. Thus to ensure that $g_\theta(I_e)$ retains the original geometry, we impose that its normal maps match those from $g_\theta(I_c)$. For each viewpoint $i$, we want to achieve:
\begin{equation}
\begin{aligned}
    N_i(I_e) &\approx N_i(I_o)\\
    \min F(g_\theta)= \sum_{i=1}^{6} &\|N_i(I_e)- N_i(I_o)\|^2
\end{aligned}
\end{equation}

Unlike optimization-based methods that attempt to modify the pre-trained weights $\theta$ to achieve the optimization objective, we operate in the latent space with fixed $\theta$. A direct solution is to replace $N_i(I_e)$ with $N_i(I_o)$ of every view $i$, which involves replacing the normal latents of the edited pipeline at every step during a single inference.

Revisiting Eq.\ref{eq2}, the edited pipeline can be formulated as:
\begin{equation}
\begin{aligned}
    g_\theta^e(T) = \prod_{t=1}^{T}g_\theta^t(I_e, N_i^t&(I_o)) = \prod_{t=1}^{T}\{ (C_i^t(I_e), \hat{N}_i^t(I_e)) \}_{i=1}^6\\
\end{aligned}
\label{eq4}
\end{equation}

where $\hat{N}_i^t(I_e) = N_i^t(I_o)$ for every step $t$ and $\prod_{t=1}^{T}$ means a step-by-step iterative process.

Nonetheless, while such manipulation effectively preserves the geometric consistency between the edited and the original normals, the alignment between the generated color images and normal maps is severely compromised. This is because at each step, the color images are combined with brand new latents from the other pipeline, resulting in their failure to align towards the polishing direction of the normals.

\begin{figure*}[t]
\centering
\includegraphics[width=1\textwidth]{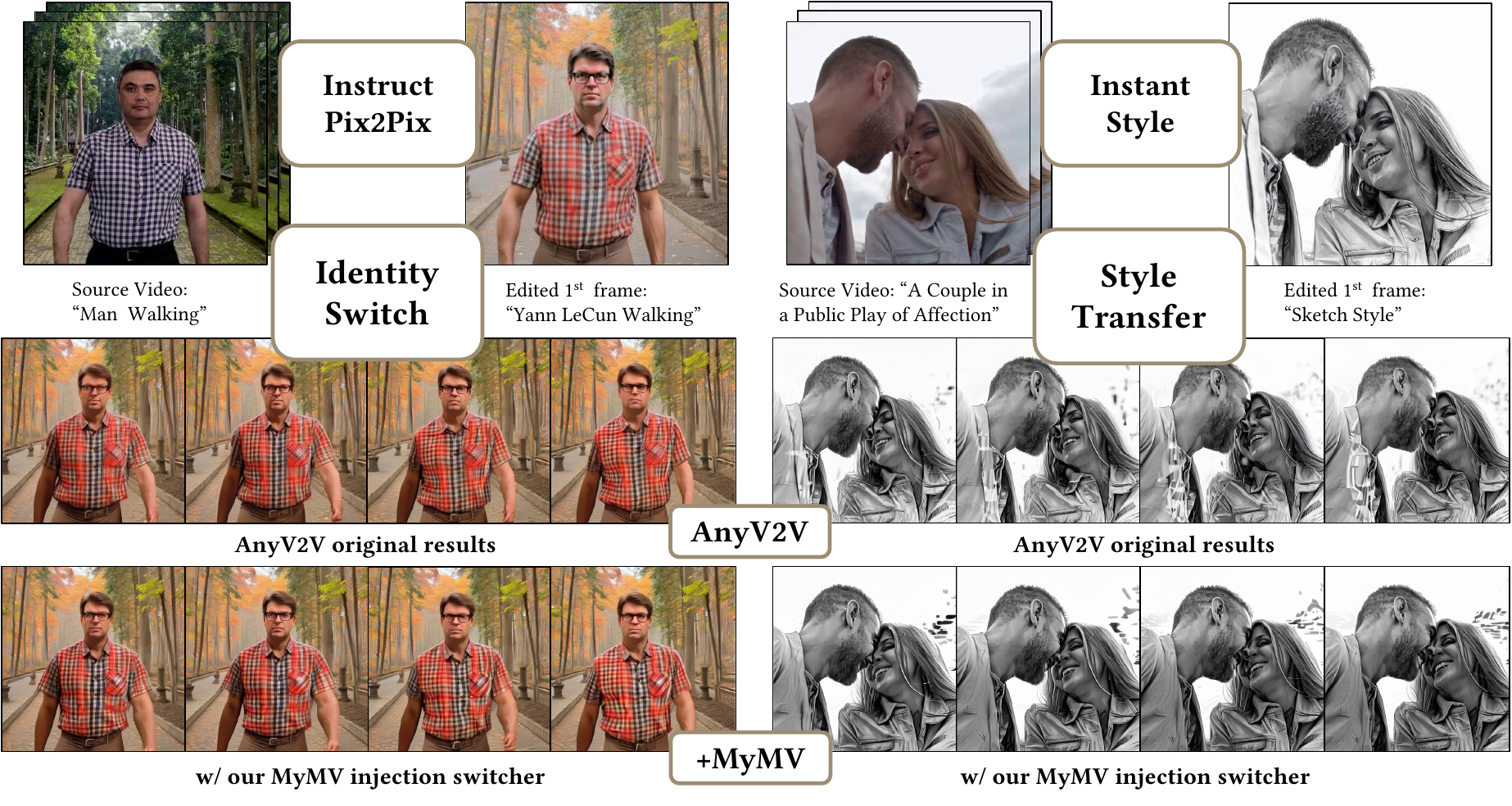} 
\caption{More application examples in video editing based on the AnyV2V, including tasks of video identity switch and video style transfer. Please zoom in for details.}
\label{moreapp}
\end{figure*}

\subsection{Injection Switcher}

To address the issue above, we design an injection switcher $s$ as a one-dimensional vector to control the extent of the original normal latents' injection. It serves as an additional input to our parallel dual-pipeline. At each inference step, the multi-view diffusion pipelines decide whether to inject the original normal latents based on the values of $s$. The Eq.\ref{eq4} can then be updated as:
\begin{equation}
\begin{aligned}
    g_\theta^e(T) = \prod_{t=1}^{T}g_\theta^t(I_e, &N_i^t(I_o), s) = \prod_{t=1}^{T}\{ (C_i^t(I_e), \hat{N}_i^t(I_e)) \}_{i=1}^6\\
    \text{where} ~\hat{N}_i^t(I_e)&=s\cdot N_i^t(I_o)+(1-s)\cdot N_i^t(I_e)
\end{aligned}
\end{equation}

We define $s$ to vary periodically with a hyperparamter $\lambda$, and conduct a grid search to determine its value. We also observe that applying it during the first several inference steps (till $t_s$) is enough to preserve the normals. The remaining steps effectively assist in aligning the edited color images and preserved normals through cross-domain self-attention. Overall, with the newly added injection switcher $s$, the cross-domain self-attention can be represented as:
\begin{equation}
\begin{aligned}
    \text{Attn}_{cd}(\tilde{Q}_{cd}, \tilde{K}_{cd}, \tilde{V}_{cd}) &= \text{softmax} \left( \frac{\tilde{Q}_{cd} \tilde{K}_{cd}^T}{\sqrt{d_k}} \right)\cdot \tilde{V}_{cd}\\
    [\tilde{Q},\tilde{K},\tilde{V}]_{cd} = \lambda \cdot [Q,K,&V]_{cn} + (1 - \lambda) \cdot [Q,K,V]_{cn'}\\\text{when}~t&\leq t_s
\end{aligned}
\end{equation}

We use $\lambda=0.5$ and $t_s=60\%~T$ as a default setting for their stable and robust results in our grid search (detailed in Sec 4.3). For some specific examples, these two parameters may be customized before inference for more ideal results. In general, larger $\lambda$ and $t_s$ tend to preserve the original geometric structure better but may sacrifice the alignment between color and normal views. Figure \ref{archi} visualizes the specific process of our whole method.

\subsection{More Applications}

It is worth noting that the modules of Make Your MoVe (abbreviated as MyMV) are not limited to the multi-view image field, but can also be extended to other diffusion-based editing tasks, like image and video editing. AnyV2V\cite{anyv2v} is an omnipotent and plug-and-play framework for any video-to-video editing tasks, leveraging a parallel I2V pipeline to replace the intermediate features. Under this setting, our injection switcher allows controlled switching between the original and edited pipeline, providing a more fine-grained feature replacement procedure and thus can generate more desirable results. Figure \ref{moreapp} showcases the polish brought by MyMV injection switcher. The facial deformation of the walking man becomes more natural and the erroneous noise generated on the couple's clothing is also properly fixed.

\section{Experiments}

In this section, we qualitatively and quantitatively compare the application of Make Your MoVe (MyMV) to two multi-view diffusion models with four editing tools against the simple combination of these methods and Style3D, demonstrating its effectiveness. Limited by space, our implementation details, additional experimental results and 360° mesh videos can be found in Appendix C and supplementary materials.

\begin{figure*}[t]
\centering
\includegraphics[width=1\textwidth]{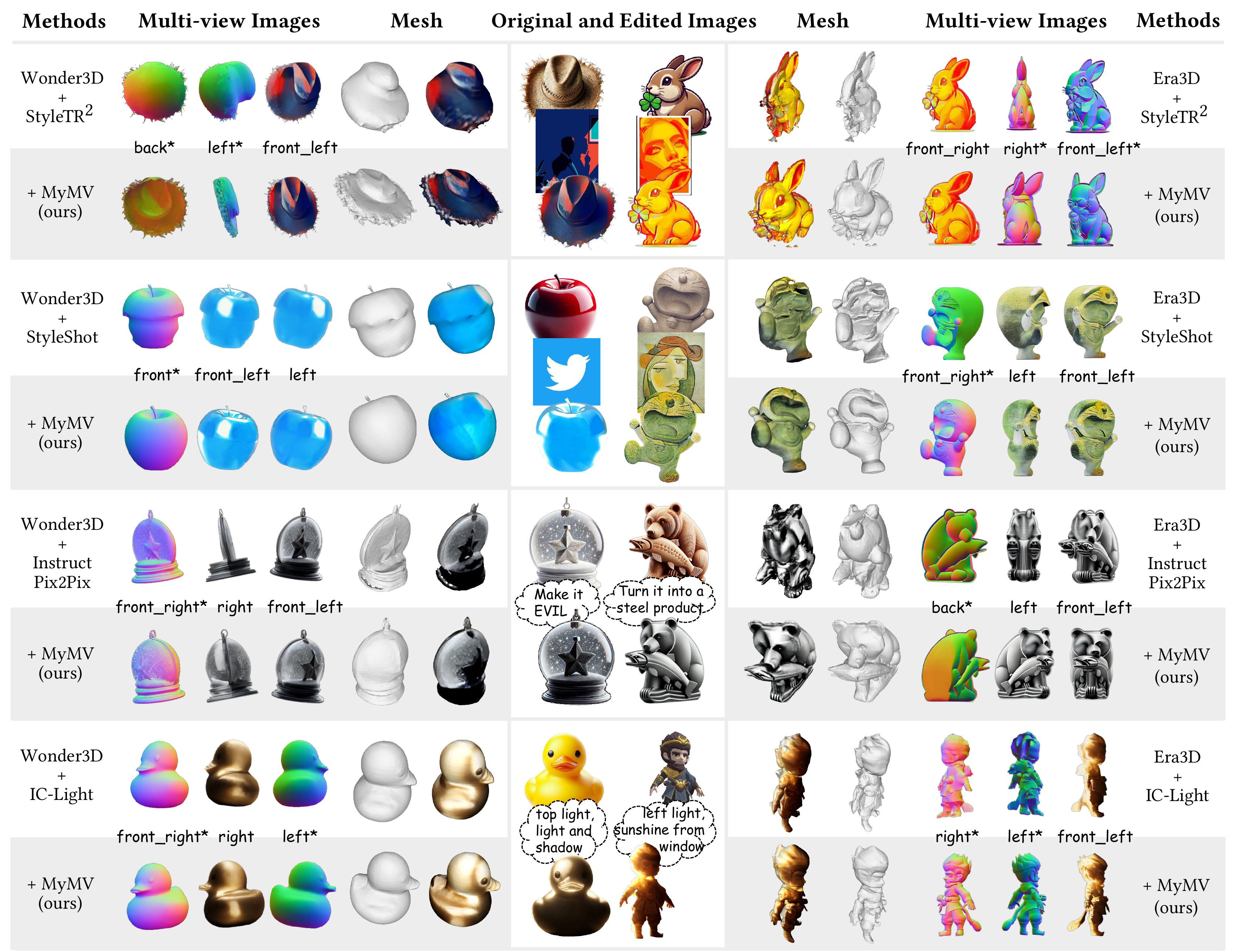} 
\caption{Qualitative comparisons between naively combining multi-view diffusions with editing tools and our \colorbox{gray!20}{MyMV}.}
\label{quanti}
\vspace{-1.3em}
\end{figure*}

\begin{figure*}[t]
\centering
\includegraphics[width=1\textwidth]{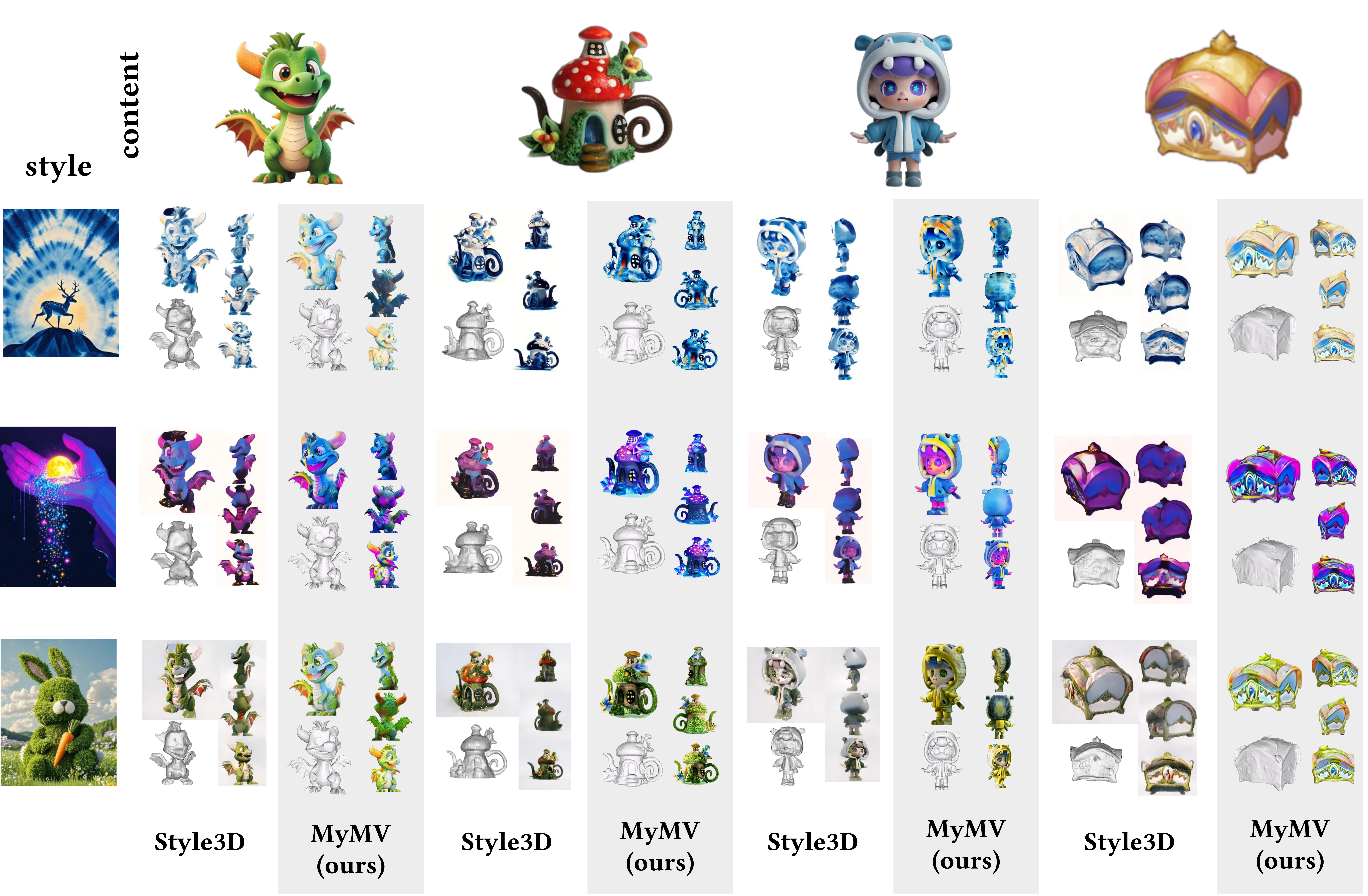} 
\caption{Qualitative comparisons between Style3D and our \colorbox{gray!20}{MyMV} using the example content and style images of Style3D.}
\label{style3d}
\end{figure*}

\subsection{Baseline and Metrics}
Wonder3D and Era3D are selected as our baselines due to their high degree of open-source availability and robust performance. We combine them with four different editing methods—StyleTR$^2$ and StyleShot (style transfer), InstructPix2Pix (prompt instruction), and IC-Light (relighting)—to demonstrate the generality and plug-and-play nature of our approach. For edits from InstructPix2Pix, we only select those not intending to affect the geometry to guarantee a fair comparison with other stylistic editing. For mesh extraction, we use the Instant-NSR from Wonder3D.

MVReward \cite{mvreward} is a novel metric for evaluating the quality of multi-view images, which can serve as an alternative to user studies as it is trained on a large RLHF dataset to predict human preferences. We evaluate each sample in the MVReward standard image-to-3D test set and those test samples provided by Wonder3D and Era3D, with random editing. Additionally, CLIP Similarity\cite{clip} is employed to measure the consistency between multi-view images.
\vspace{-0.5em}
\subsection{Evaluation}

It is an intuitive way that we compare MyMV with \emph{naive} combinations of 2D editing methods and multi-view diffusion baselines. Notably, we are the first to argue that multimodal editing methods should first compare with [image editing + image-to-modality] combinations, which are neglected by previous work like Style3D ([image editing + image-to-3D]) and AnyV2V ([image editing + image-to-video]). Such comparison is essential since it illustrates the necessity of the editing task rather than simply relying on existing tools. Furthermore, we also compare with Style3D as it is a SOTA method in 3D editing based on multi-view images. 

\noindent\textbf{Qualitative Results.} Figure \ref{quanti} presents several representative visual comparisons with \emph{naive} combinations, highlighting that our MyMV generates better multi-view images and higher mesh quality across all cases (* means normal map). MyMV corrected geometric errors in the \emph{Hat}, \emph{Crystal Ball}, and \emph{Rabbit}; filled in missing geometry for the \emph{Apple} and \emph{Duck}; resolved the multiple-head issue of the \emph{Bear}; and refined the geometry of the \emph{Doraemon} and \emph{Monkey King}.

Due to Style3D not having open-source code, we directly use the qualitative results from its original paper and compare them with ours in Figure \ref{style3d}. It is evident that our method generates more fine-grained mesh results and consistently preserves geometry across different styles for the same content. Note that there is a slight discrepancy between our camera system and Style3D, especially for the \emph{Box}, but this does not affect the visual quality comparison. The aesthetic quality of the style transfer is not attributed to our method, since we utilize existing 2D editing tools, and the appearance changes are mainly determined by them. Our focus lies in preserving geometry and ensuring alignment between color images and normal maps, which is reflected in the quality of the final generated meshes.

\noindent\textbf{Quantitative Comparison.} We compare our method with \emph{naive} combinations on MVReward, CLIP Similarity, and Time in Table \ref{table}. It demonstrates that our MyMV consistently improves the alignment between edited multi-view images and human preferences, as well as cross-view consistency, with acceptable additional time overhead, which validates the effectiveness and efficiency of our approach. The quantitative mesh evaluation and comparisons with non-mesh based methods can be found in Appendix C.

\noindent\textbf{User Study.} We conduct a user study to further evaluate our methods. We provide users with visual comparisons between the \emph{naive} combinations and ours, as well as between Style3D and ours. They are asked to choose the better one separately for three aspects: geometric plausibility, cross-view consistency, and mesh quality. The win rates are reported in Table \ref{human}, our method robustly outperforms the \emph{naive} combinations and Style3D, underscoring its superiority.

\renewcommand{\arraystretch}{1.0}
\begin{table}[t]
\small
\resizebox{0.48\textwidth}{!}{
\begin{tabular}{ccc>{\columncolor[gray]{0.9}}c}
\toprule
      Win Rate                 & naive & tie    & \cellcolor{white}MyMV   \\ \midrule
geometric plausibility & 2.4\%              & 6.5\%  & 91.1\% \\
cross-view consistency & 8.2\%              & 25.7\% & 66.1\% \\
mesh quality           & 5.1\%              & 4.5\%  & 90.4\% \\ \bottomrule
\end{tabular}
}

\vspace{0.5em}

\resizebox{0.48\textwidth}{!}{
\begin{tabular}{ccc>{\columncolor[gray]{0.9}}c}
\toprule
      Win Rate                 & Style3D & tie    & \cellcolor{white}MyMV   \\ \midrule
geometric plausibility & 16.7\%              & 8.3\%  & 75.0\% \\
cross-view consistency & 8.3\%              & 25.0\% & 66.7\% \\
mesh quality           & 8.3\%              & 0\%  & 91.7\% \\ \bottomrule
\end{tabular}
}

\caption{User study between naive combinations and our \colorbox{gray!20}{MyMV}, as well as Style3D and our \colorbox{gray!20}{MyMV}.}
\label{human}
\vspace{-1em}
\end{table}
\subsection{Ablation Study}
\renewcommand{\arraystretch}{1.6}
\begin{table*}[]
\centering
\resizebox{\textwidth}{!}{
\begin{tabular}{rcccrccc}
\toprule
Combinations             & MVReward & CLIP Similarity & Time(s) & Combinations             & MVReward & CLIP Similarity & Time(s)\\ \midrule
Wonder3D + StyleTR$^2$       &   0.71       &         0.78           &   3.3   &  Era3D + StyleTR$^2$          &   1.47       &             0.82        &   14.6   \\
\rowcolor[gray]{0.9}+MyMV (ours)&   \textbf{0.82}       &         0.77            &   6.7  &  +MyMV (ours) &   \textbf{1.78}       &           \textbf{0.85}          &   29.4   \\ \hline

Wonder3D + StyleShot       &   0.74       &        0.75             &   39.1   &  Era3D + StyleShot         &   1.01                 &     0.81      &   50.6   \\
\rowcolor[gray]{0.9}+MyMV (ours)&   \textbf{0.92}       &     \textbf{0.77}                 &   42.3  & +MyMV (ours)&   \textbf{1.69}      &       \textbf{0.85}               &   65.3   \\ \hline

Wonder3D + InstructPix2Pix       &   1.31       &      0.79                &   4.8   &  Era3D + InstructPix2Pix         &   1.49                  &    0.76       &   15.5   \\
\rowcolor[gray]{0.9}+MyMV (ours)&   \textbf{1.43}      &       0.79               &   8.0  & +MyMV (ours)&   \textbf{1.80}       &        \textbf{0.80}              &   30.3   \\ \hline

Wonder3D + IC-Light       &   1.07       &       0.76               &   12.8   &  Era3D + IC-Light         &   0.94       &                0.76      &   24.2   \\
\rowcolor[gray]{0.9}+MyMV (ours)&   \textbf{1.19}       &      \textbf{0.82}               &   15.9  & +MyMV (ours)&   \textbf{1.25}       &    \textbf{0.83}                 &   39.0   \\ \hline

\end{tabular}
}
\caption{Quantitative comparisons between naively combining multi-view diffusions with editing tools and our \colorbox{gray!20}{MyMV}.}
\label{table}
\vspace{-1.8em}
\end{table*}
\begin{figure}[t]
\centering
\includegraphics[width=0.5\textwidth]{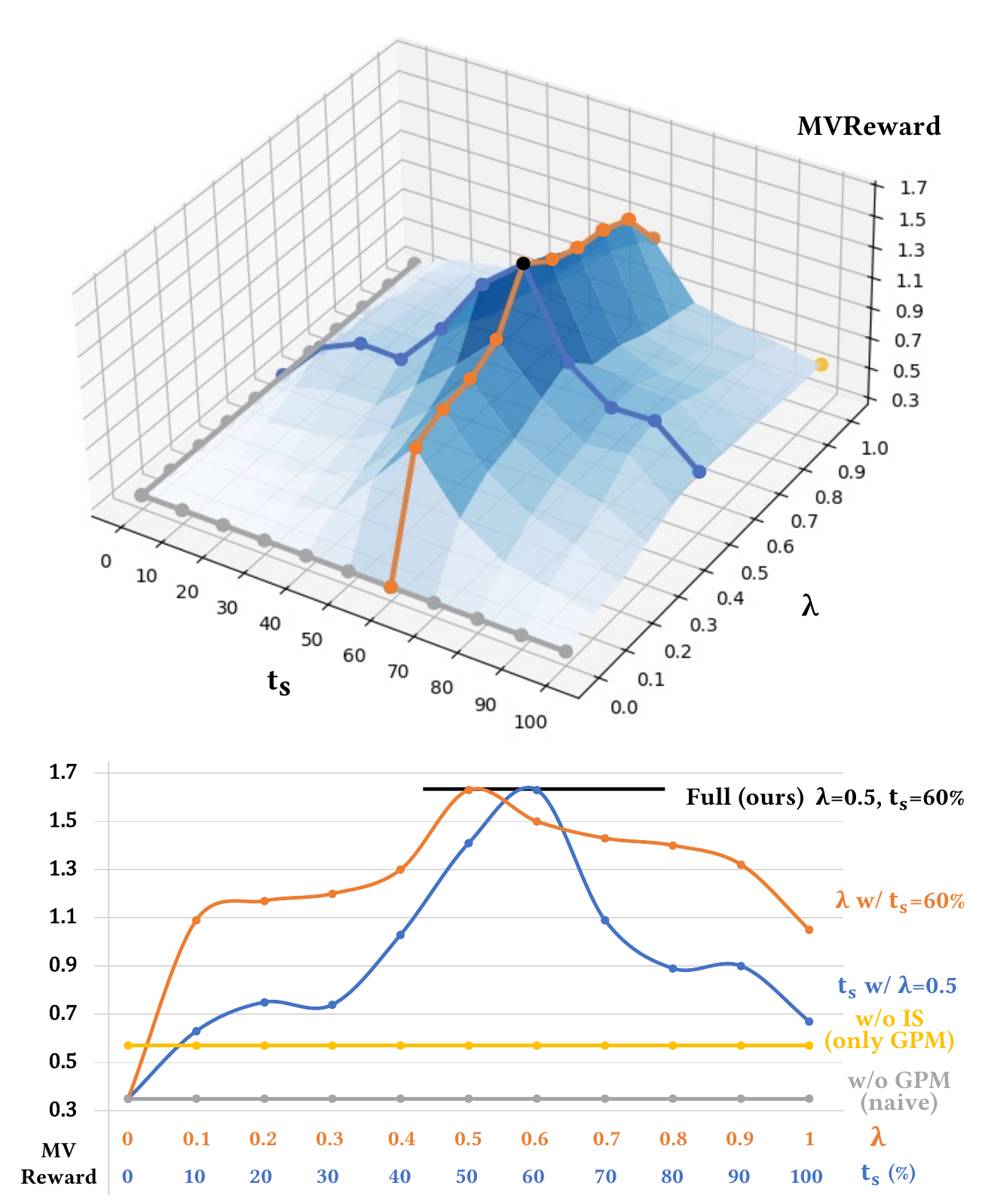} 
\caption{Quantitative ablation experiment (grid search) on our geometry preservation module (GPM), injection switcher (IS), and two hyperparameters $\lambda$ and $t_s$.}
\label{abla2}
\vspace{-1em}
\end{figure}
To examine the contribution of our proposed modules, we ablate on the geometry preservation module, injection switcher, and two hyperparameters. A grid search is conducted on 100 samples from the MVReward standard test set, with the results visualized in Figure \ref{abla2}. The upper is a 3D surface plot of the full ablation space, while the lower highlights a comparison of several key configurations. When either $\lambda$ or $t_s$ is set to 0, our method is disabled and degenerates into the \emph{naive} combination (gray line). When $\lambda = 1.0$ and $t_s = 100$, the impact of the injection switcher is effectively removed, leaving only the geometry preservation module enabled (yellow dot/line). The orange and blue lines represent results where $\lambda$ and $t_s$ are set to their respective optimal values. Their intersection point (black dot), which corresponds to the peak of the 3D surface, denotes the default setting used in our method.

\begin{figure}[t]
\centering
\includegraphics[width=0.48\textwidth]{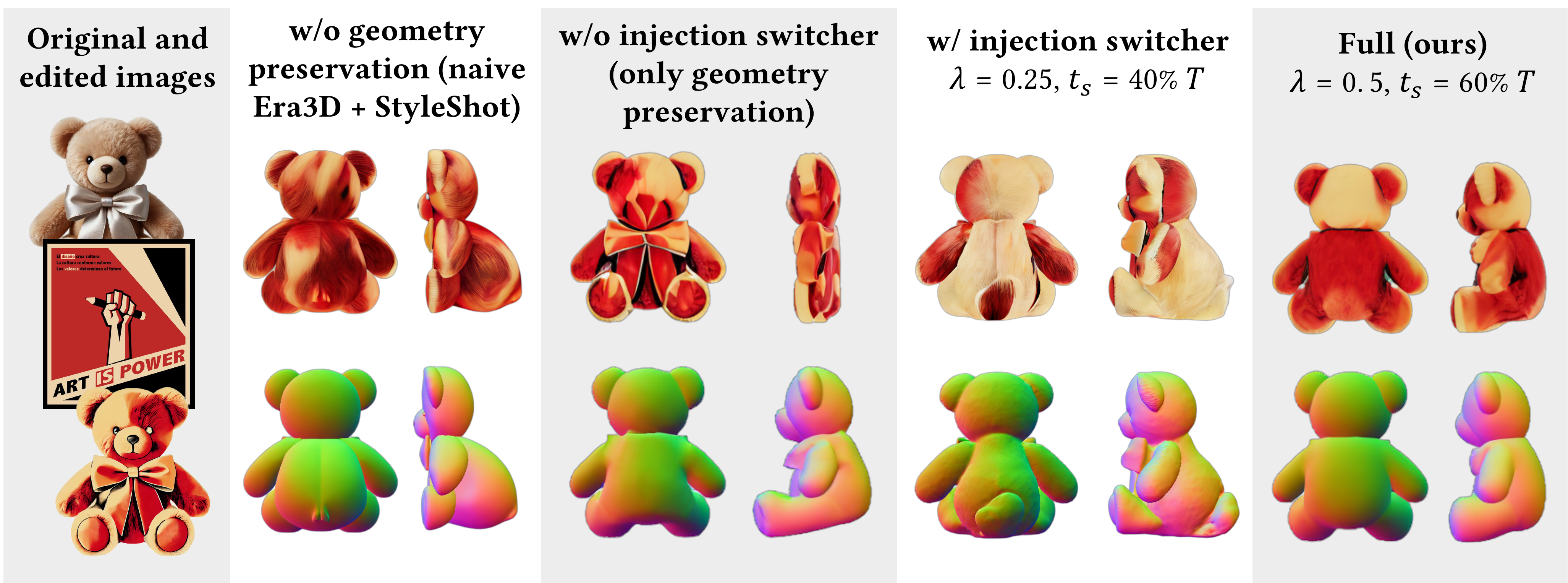} 
\caption{Qualitative ablation experiment (representative case) on our geometry preservation module, injection switcher, and two hyperparameters $\lambda$ and $t_s$.}
\label{abla1}
\vspace{-1em}
\end{figure}

We further select one representative ablation case in Figure \ref{abla1}. From the changes of the stylized teddy bear, we can see that using only our geometry preservation module can correct distorted normal maps, but brings misalignment between the color and normal views which is addressed by our injection switcher. For the hyperparameters $\lambda$ and $t_s$, smaller values lead to a greater influence of the distorted geometry, similar to interpolation between the original and edited normals, and our default setting proves to be a sweet spot. All the findings are consistent with Sec. 3.4 analysis.

\section{Conclusion}

In this paper, we fully analyze and address the challenges in adapting multi-view diffusion models to external editing. To mitigate the potential information loss or geometry blurring introduced by existing 2D editing tools, we propose a geometry preservation module to retain the original geometry of the generated 3D assets. Additionally, we introduce an injection switcher to precisely control the degree of supervision from the original normals, thereby enhancing the alignment of the edited color and normal views. Extensive experimental results support our contributions in providing a tuning-free, plug-and-play scheme to improve the multi-view consistency and mesh quality of edited 3D assets, compared with naive combinations and a SOTA 3D appearance editing method. For a further and more detailed discussion on our social impact, ethical concerns, limitations, and future work, please refer to Appendix D.

\bibliographystyle{ACM-Reference-Format}
\bibliography{sample-base}

\end{document}